\documentclass[letterpaper]{article}
\usepackage{uai2020}
\usepackage[margin=1in]{geometry}

\usepackage{graphicx}
\usepackage{subcaption}

\usepackage{booktabs}
\usepackage{mathtools}
\usepackage{multirow}

\usepackage{amsmath}
\usepackage{amssymb}

\newcommand{\data}{\mathcal{D}}

\usepackage{tikz}
\usetikzlibrary{shapes,decorations,arrows,calc,arrows.meta,fit,positioning}
\tikzset{
    -Latex,auto,node distance =1 cm and 1 cm,semithick,
    state/.style ={circle, draw, minimum width = 0.7 cm},
    data/.style ={circle, draw, fill=black!10, minimum width = 0.7 cm},
    constant/.style ={rectangle, draw, minimum width = 0.7 cm}
}

\usepackage[hyphens]{url}
 \usepackage[hidelinks]{hyperref}

\usepackage{natbib}

\title{GPIRT: A Gaussian Process Model for Item Response Theory}

\author{{\bf JBrandon Duck-Mayr} \\
Washington University in St.\ Louis \\
\texttt{j.duck-mayr@wustl.edu} \\
\And
{\bf Roman Garnett} \\
Washington University in St.\ Louis \\
\texttt{garnett@wustl.edu} \\
\And
{\bf Jacob M. Montgomery} \\
Washington University in St.\ Louis \\
\texttt{jacob.montgomery@wustl.edu}
}

\begin{document}

\maketitle

\begin{abstract}
    The goal of item response theoretic (IRT) models is to provide estimates of latent traits from binary observed indicators and at the same time to learn the item response functions (IRFs) that map from latent trait to observed response.
    However, in many cases observed behavior can deviate significantly from the parametric assumptions of traditional IRT models.
    Nonparametric IRT models overcome these challenges by relaxing assumptions about the form of the IRFs, but standard tools are unable to simultaneously estimate flexible IRFs and recover ability estimates for respondents.
    We propose a Bayesian nonparametric model that solves this problem by placing Gaussian process priors on the latent functions defining the IRFs.
    This allows us to simultaneously relax assumptions about the shape of the IRFs while preserving the ability to estimate latent traits.
    This in turn allows us to easily extend the model to further tasks such as active learning.
    GPIRT therefore provides a simple and intuitive solution to several longstanding problems in the IRT literature.
\end{abstract}

\section{INTRODUCTION} 

\emph{Item response theory} (IRT) \citep{rasch1960studies, Lord:1968} is a widely used framework for estimating latent traits across many application domains, including educational testing \citep[][]{Lord:1980},  psychometrics \citep{Embretson:2000},  political science \citep{MartinQuinn:2002, Clinton:2004}, and more.  Like other dimensionality reduction techniques, the goal is to take a large set of observed features and map them to a low-dimensional representation maintaining latent structure.  A prototypical setting is assigning latent ability scores to students answering test questions.  The observed responses are typically binary or
categorical, making standard factor analysis inappropriate.
Further, in addition to estimating latent scores, we also seek to simultaneously
estimate the mapping between latent positions and observed outcomes.
These mappings are themselves of interest to applied scholars and enable downstream tasks such as  optimal test construction 
or active learning procedures like computerized adaptive testing (CAT) \citep{Weiss:1982, vanderLinden:2010}. We refer to these probabilistic mappings from latent ability to predicted outcome
as \emph{item response functions} (IRFs). 

Common IRT models are parametric, typically estimating a sigmoidal
mapping between the latent space and a binary outcome. Problems arise,
however, when respondent behavior deviates from the assumed parametric
representations.  For instance, nearly all models assume a
\emph{monotonic} relationship between a respondent's position on the
latent scale $\theta$ and each observed response $y$, which can be
violated in settings such as personality measurement
\citep[e.g.,][]{Meijer:2004}.  A further problem is
\emph{non-saturation,} where the probability of observing either
outcome never fully approaches zero or one (e.g., guessing on multiple-choice tests). A more subtle problem occurs
when there is \emph{asymmetry} in the IRFs. Tests for cognitive
 abilities require higher levels of complex thinking
for success such that the shape of the IRF for respondents with low
values of $\theta$ do not mirror the shape for individuals with high
values \citep{Lee:2018}.

Numerous parametric models have been proposed to accommodate these and other
irregularities. A complete inventory of IRT-like models would be
prohibitively long, but notable examples include: (i) the generalized graded
unfolding model (GGUM) \citep{Robertsetal:2000}, which allows for
(symmetric) non-monotonic IRFs; (ii) the three- (3PL)
\citep{birnbaum1968some} and four-parameter (4PL) \citep{Barton:1981}
logistic models that relaxed saturation assumptions by including
``guessing'' and ``carelessness'' parameters; and (iii) Samejima's (\citeyear{Samejima:2000}) 
logistic positive exponential family models (LPEF), which allow for
asymmetric IRFs.

A related literature approaches these problems from a
nonparametric framework \citep[see][]{Sijtsma:1998}.  Mokken models \citep{Mokken:1971, Mokken:1982}
relax functional form assumptions for the IRFs but require 
 monotonicity  \citep{Sijtsma:2002}, which is also required for other nonparametric techniques \citep[e.g.,][]{Poole:2000}.  Ramsay (\citeyear{Ramsay:1991}) proposes a model based on kernel smoothing. However, standard nonparametric methods are
 unsatisfactory since the IRFs and $\theta$ parameters
cannot be estimated simultaneously.  A standard approach for kernel-based IRT models, for instance, is to smooth
over the rank ordering of respondents in terms of their raw scores on
the test. Thus, while the IRFs are flexible, the smoothing
occurs not over the latent parameters $\theta$, but over
``reasonable estimates'' of $\theta$ constructed from $y$
\citep[p. 4]{Mazza:2012}.  Further, since inference is not based on the true
likelihood, these models extend poorly to
downstream tasks such as active learning.

In this article, we propose a Bayesian nonparametric IRT model based on
 Gaussian process (GP) regression \citep{RW:2006}. As we demonstrate, our model has several clear advantages.
First, similar to existing kernel smoothing methods, the Gaussian process IRT model (GPIRT) can in principal recover any
smooth IRF with minimal assumptions. However, adopting the Bayesian framework facilitates building smoothed IRFs in a more coherent manner based on the actual latent parameters rather than proxies constructed from $y$. Second, GPIRT is a direct extension of Bayesian parametric IRT models
\citep{albert1992bayesian, albert1993bayesian}. Indeed, many parametric Bayesian models in the literature 
can be viewed as a special case of our more general framework. Finally,  it is simple to extend the model for downstream tasks such as item diagnostics, test construction, and computerized adaptive testing.

Below we use GPIRT to generate more accurate estimates of latent
ability parameters while simultaneously capturing smooth IRFs that are
non-monotonic, non-saturating, and asymmetric.  We demonstrate
its flexibility with real-world data, including voting records from
the US Congress and responses to a personality inventory measuring
narcissism, and contrast the behavior of GPIRT with parametric
baselines. We also demonstrate how to use GPIRT to implement
seamless active learning by maximizing mutual information to rapidly
determine a new respondent's latent score from their responses to
adaptively chosen items. This enables dynamic test construction in the
tradition of CAT, and we will show it performs admirably on real-world
data.

\section{OVERVIEW OF IRT}

We start by presenting the groundwork for IRT models. We then introduce the GPIRT, describe an MCMC
sampling algorithm, extend the model for active learning,  and contrast it with prominent nonparametric IRT approaches in the literature. We conclude with two applications. We restrict our discussion to the unidimensional case, where items are
assumed to load on a single underlying latent dimension; however, our model
would extend to the multidimensional setting. The multidimensional case has
 been explored previously in the nonparametric IRT setting \citep[e.g.,][]{bartolucci2017nonparametric}, but it is not as common
 in practice. We will also focus on the binary response case, which is the canonical IRT setting. However, categorical responses can be handled via standard extensions to the model below.

\subsection{OBSERVATION MODEL}

Assume we have $n$ binary-response items and $m$ respondents who have each answered
some subset of items. We will encode the response of respondent $j$ on
item $i$ as $y_{ij}\in \{-1, 1\}$, where 1 encodes a positive (or correct) response and $-1$ denotes an
negative (or incorrect) response.  This presentation is in keeping with GP classification models 
and deviates trivially from standard IRT presentations where we typically take $y \in \{0, 1\}$.
 
IRT assumes a simple probabilistic model for $y$ according to (i) a latent score associated with each respondent and (ii) an item-specific mapping between latent score and the probability of observing $y_{ij}=1$. 
Let $\Theta$ be some latent space (here we take $\Theta = \mathbb{R}$) in which we wish to embed respondents. 
We assume each respondent $j$ has some unknown location in this space, $\theta_j$ and that that each item $i$ has an associated latent function $f_i\colon \Theta \to \mathbb{R}$ that gives rise to responses via a sigmoidal link function. 
Namely, we assume the probability that respondent $j$ answers item $i$ positively is
\begin{equation}
    \label{eq:basic}
  \Pr(y_{ij}=1 \mid \theta_j, f_i) = \sigma\bigl(f_i(\theta_j)\bigr),
\end{equation}
where $\sigma\colon \mathbb{R} \to (0, 1)$ is a monotonic sigmoidal ``squashing'' function such as the logistic or standard normal CDF (inverse probit).
We will further adopt the standard assumption that multiple responses across a set of items/respondents are conditionally independent given the latent scores and latent functions. For a set of observed responses $\{y_{ij}\}$, this results in the likelihood:
\begin{equation}
\label{eq:likelihood}
  \Pr\bigl( \{y_{ij}\} \mid \{f_i\}, \{\theta_j\} \bigr)
  = 
  \prod_i \prod_j \Pr(y_{ij} \mid \theta_j, f_i),
\end{equation}
where the product extends over the set of responses.

Interpreted as a function of $\theta_j$, the marginal probability of a positive response to item $i$, Equation \eqref{eq:basic} is known as the $i$th \emph{item response function} (IRF). The goal of IRT is to jointly estimate the respondents' latent locations $\{\theta_j\}$ and the IRFs from some set of training observations $\{y_{ij}\}$. This may seem daunting since we are in essence trying to simultaneously learn a latent embedding \emph{and} the mappings $\{f_i\}$ using only observed responses.  The problem is tractable if we assume that the latent functions $\{f_i\}$ are smooth over $\Theta$. In that case, the fact that each respondent corresponds to a \emph{single} latent location shared by all the IRFs gives us hope that we can find an embedding that places respondents with correlated responses in neighboring regions of the latent space.

\subsection{PARAMETRIC IRT MODELS AND CAT}

\begin{figure}[!t]
    \centering
    \begin{subfigure}[b]{0.9\linewidth}
        \includegraphics[width=\textwidth]{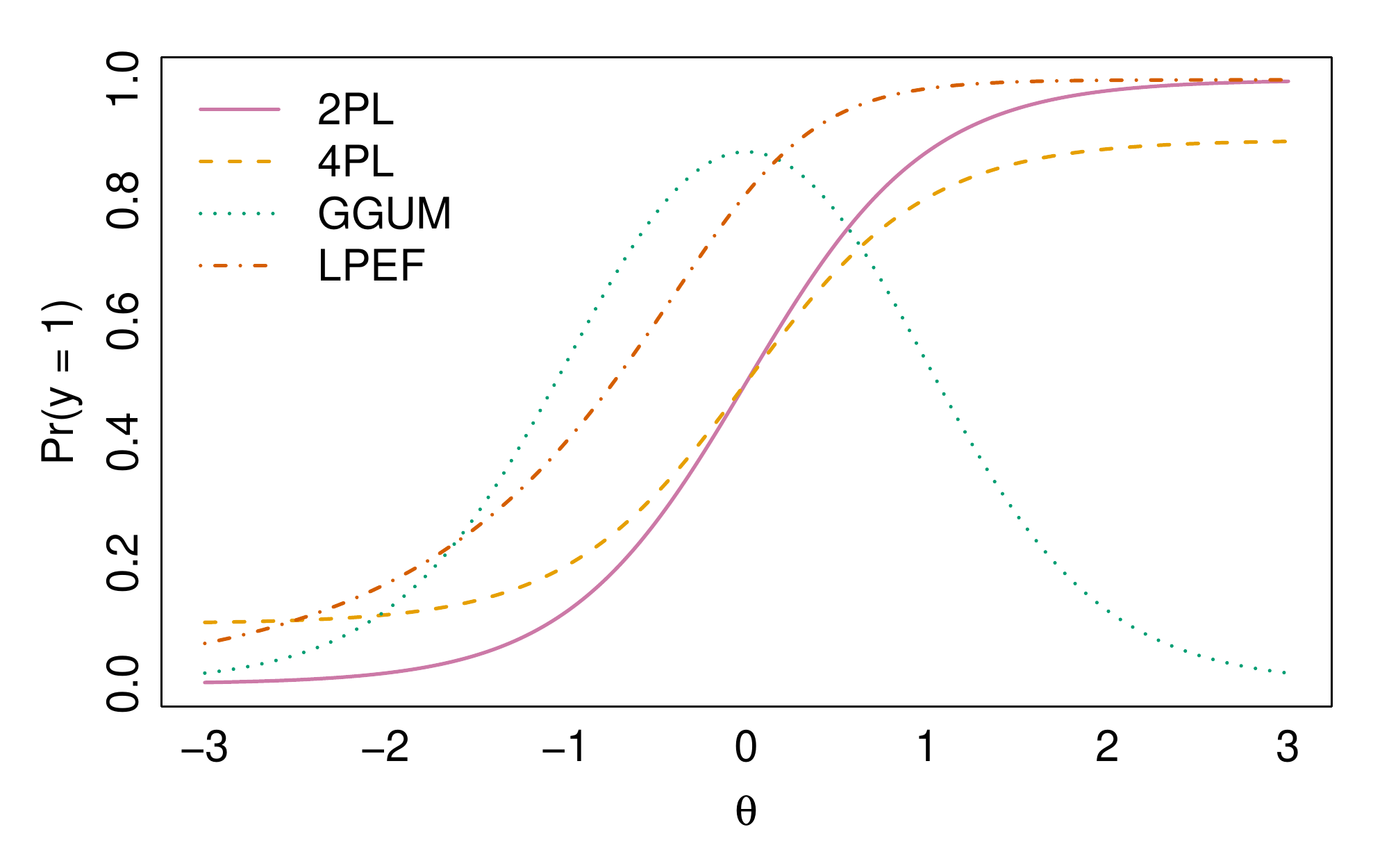}
        \caption{IRFs for standard IRT models}
        \label{fig:traditional-irf-examples}
    \end{subfigure}
    \begin{subfigure}[b]{0.9\linewidth}
        \includegraphics[width=\textwidth]{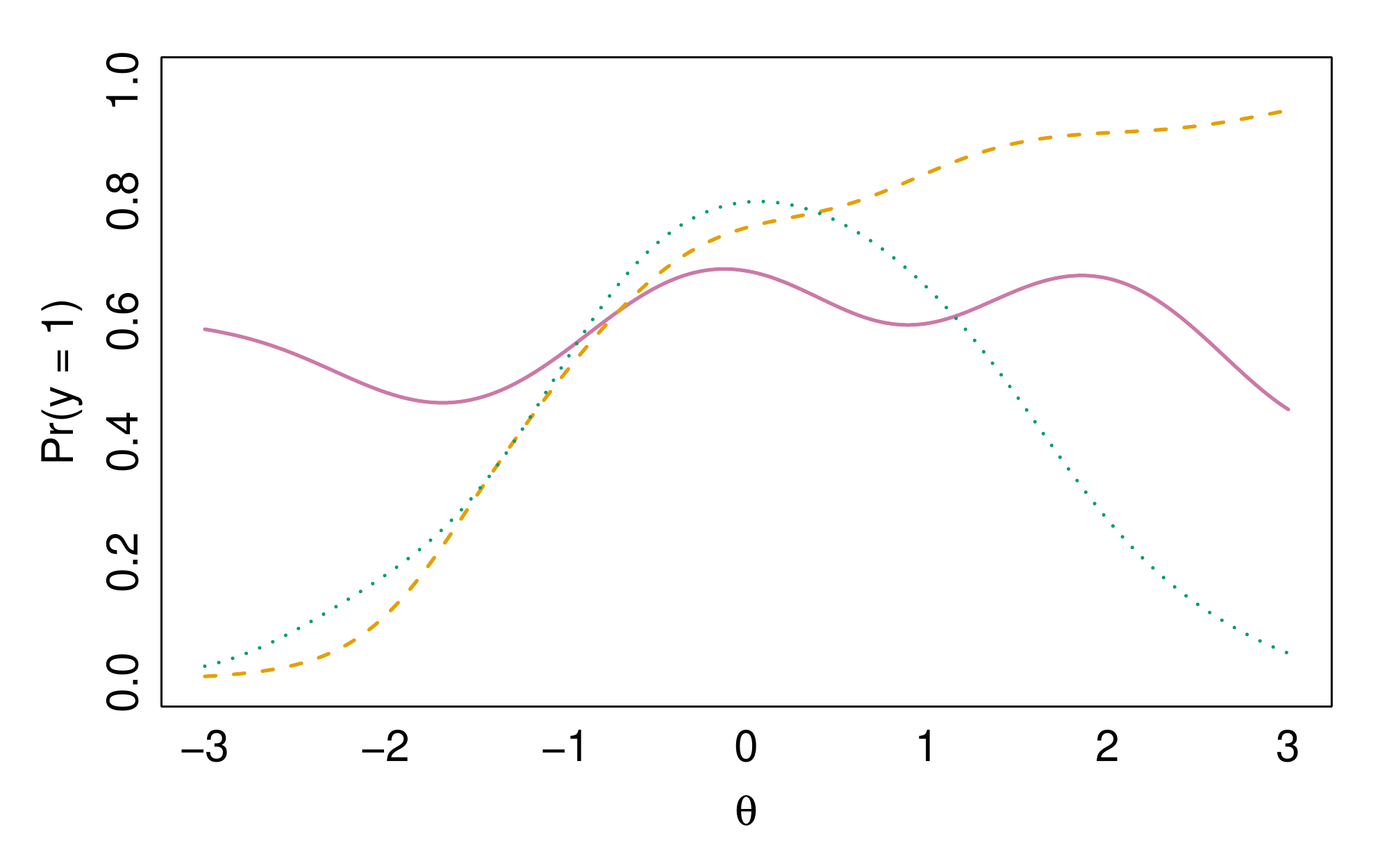}
        \caption{IRFs for GPIRT}
        \label{fig:gp-irf-examples}
    \end{subfigure}
    \caption{\footnotesize Example IRFs for the 2PL, 4PL, GGUM, and LPEF models compared to IRFs for GP latent functions.}
    \label{fig:example-irfs}
\end{figure}

In the two-parameter logistic model (2PL) \citep{birnbaum1968some}, $\sigma$ is taken to be the logistic function and we assume a linear parametric form for the latent functions:
\[
   f_i(\theta_j) = \beta_{i0} + \beta_{i1} \theta_j.
\]
By convention, $\theta_j$ is referred to as the \textit{ability} parameter for respondent $j$. 
These parametric assumptions are quite restrictive and fail in numerous settings. As a consequence, a wide array of alternative parametric forms for the $\{f_i\}$ have been proposed, including the GGUM, 3PL, 4PL, and LPEF models discussed above. Example IRFs for each are shown in Figure 1(a).

Critically, the IRFs are themselves often of interest to researchers.  Sometimes, as in our analysis of the US Congress below, the shape of the IRF provides important insights about a specific item. In other cases, such as measuring personality traits, IRFs are used for test construction where the goal is to choose a set of items (i.e., a test in the sense of an examination) that will
reveal as much as possible about the target population. This includes computerized adaptive testing, where items are chosen dynamically during test administration \citep{Weiss:1982}.  CAT is widely used in educational testing \citep{vanderLinden:2010}, psychology \citep{Waller:1989}, and survey research \citep{Montgomery:forthcoming}. 

A number of CAT algorithms have been proposed in the literature.
Of particular relevance here are the maximum expected Kullback--Leibler (MEKL) criterion
\citep{Chang:1996} and its Bayesian variant, the maximum posterior weighted Kullback-Leibler (MPWKL) criterion \citep{vanderLinden:1998},
which is equivalent to the mutual information between an unknown response and the latent ability score.%
\footnote{In the CAT literature the first appearance of MEKL, which might sound equivalent to mutual information, was defined in a somewhat idiosyncratic manner.}
These methods seek to accelerate the convergence of our estimate of $\theta$ to the true unknown posterior by maximizing the
expected information gain from each item selected from the larger inventory of potential items.
MPWKL can be directly extended to our GPIRT model.
It is worth noting that all of the above approaches for CAT assume a pre-calibrated parametric IRT model.
To our knowledge \citet{Xu:2006} present the only nonparametric IRT CAT algorithm.

\section{GPIRT}

Our proposed model, the Gaussian process IRT (GPIRT) model, extends standard models by placing Gaussian process priors on the latent functions $\{f\}$.  We then perform joint inference over latent functions and ability scores.
This procedure allows us to quantify and propagate our typical lack of certainty about the shape of the IRFs, and allows us to readily adopt the advances in Gaussian process modeling and inference seen over the past decade for estimation and active learning.

Although constructing and performing inference in this model will be straightforward with existing machinery,
GPIRT provides a simple and intuitive solution to several longstanding problems in the IRT literature.
The model makes no assumptions about the shape of IRFs beyond smoothness over the latent score $\theta$.
In principle, this allows the resulting estimated IRFs to take the shape of any smooth function (assuming suitable flexibility in the choice of covariance function) and may be non-monotonic, non-saturating, non-symmetric, or all of the above. 
IRFs derived from draws from a GP prior are shown in Figure 1(b).
Further, adopting the Bayesian framework allows us to simultaneously estimate these flexible IRFs and the latent scores via a sampling approach we will describe below.
This removes potential bias in IRF estimation from error in score estimation
as well as relaxing the monotonicity assumption of Mokken-like models.
Finally, the construction of the model facilitates item selection in either a manual or adaptive testing setting by adopting established ideas from Bayesian experimental design.

The most critical innovation in the GPIRT model is that rather than assuming that the functions $\{f_i\}$ belong to a specific parametric family, we place an independent Gaussian process prior distribution on each:
\begin{equation}
  p(f_i) = GP(f_i; \mu, K),
\end{equation}
where $\mu$ is a shared prior mean function that can be chosen according to prior beliefs,
and $K$ is a covariance function between respondents' latent traits. We could choose the covariance function as we see fit; however,
for this investigation we took the pervasive squared exponential covariance function with unit length scale:
\begin{equation}
    K(\theta, \theta') = \exp\bigl(-\tfrac{1}{2} ( \theta - \theta')^2 \bigr).
    \label{sqdexp}
\end{equation}
Note that we will also be taking a standard normal prior on the latent scores $\{\theta\}$ and thus the unit length scale is compatible with our expected range of
latent scores.

For the mean function, $\mu \equiv 0$ is the most agnostic (corresponding to a prior IRF assigning
50\% probability to a positive response regardless of location),
but does not leverage prior knowledge about IRF profiles.
Another reasonable choice is a linear prior mean $ \mu(\theta) = \beta_{i0} + \beta_{i1} \theta,$ which is most appropriate in a context where monotonic IRFs are expected, but where we prefer not to impose \emph{strict} linearity.
Note that with a linear mean we can recover standard two-parameter IRT models by selecting $\sigma$ appropriately and taking
$K$ to be a linear covariance. GPIRT therefore provides a nonparametric generalization where each item is explained by a latent linear trend with smooth nonlinear deviations.

To complete the model, we place an independent normal prior over the respondents' latent scores $\{ \theta \}$:
\[
  p\bigl(\{\theta_j\}\bigr) = \textstyle \prod_j \phi(\theta_j),
\]
where $\phi$ is the standard normal PDF. We also place independent normal priors over the coefficients $\{\beta\}$ for linear or higher-order polynomial mean functions as necessary.

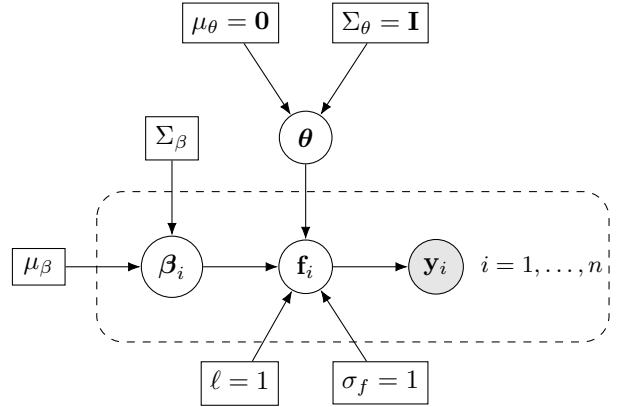
\begin{figure}[!t]

\begin{center}
    \begin{tikzpicture}
        \node[state]    (t)          at (0, 0)                {$\boldsymbol\theta$};  
        \node[state]    (f)   [below       =     of t]        {$\mathbf{f}_i$}; 
        \node[state]    (b)   [left        =     of f]        {$\boldsymbol\beta_i$};       
        \node[data]     (y)   [right       =     of f]        {$\mathbf{y}_i$};       
        \node[constant] (m_t) [above left  = 1cm and 1mm of t] {$\mu_\theta = \mathbf{0}$};
        \node[constant] (s_t) [above right = 1cm and 1mm of t] {$\Sigma_\theta = \mathbf{I}$};
        \node[constant] (m_b) [left        =     of b]         {$\mu_\beta$};
        \node[constant] (s_b) [above       =     of b]         {$\Sigma_\beta$};
        \node[constant] (sf)  [below left  = 1cm and 1mm of f] {$\ell\vphantom{_f} = 1$};
        \node[constant] (ls)  [below right = 1cm and 1mm of f] {$\sigma_f = 1$};
        \path (t)   edge (f);
        \path (b)   edge (f);
        \path (f)   edge (y);
        \path (m_t) edge (t);
        \path (s_t) edge (t);
        \path (m_b) edge (b);
        \path (s_b) edge (b);
        \path (sf)  edge (f);
        \path (ls)  edge (f);
        \draw[dashed,rounded corners=10]($(b) + (-1,1)$)rectangle($(y) +(2.3,-1)$);
        \node[right = 1mm of y] {{\footnotesize$i = 1, \ldots, n$}};
    \end{tikzpicture}
\end{center}
    \caption{\footnotesize Graphical representation of GPIRT.}
    \label{fig:plate}
\end{figure}

A graphical representation of the model is shown in Figure~\ref{fig:plate}.  Here, $\mu_\theta$ and $\Sigma_\theta$ are the mean and covariance of the prior on $\boldsymbol\theta$.  Similarly, $\mu_\beta$ and $\Sigma_\beta$ are the mean and covariance of the prior on $\boldsymbol\beta$. Finally,
$\sigma_f$ and $\ell\vphantom{_f}$ are the scale factor and length scale of $K$ respectively.

\subsection{ESTIMATION}

Assume we have a set of responses $\mathcal{D} = \{y_{ij}\}$. We wish to estimate the posterior over the latent functions $\{f\}$ and latent scores $\{\theta\}$, $p\bigl(\{f\}, \{\theta\} \mid \mathcal{D}\bigr)$. 
Since the posterior is analytically intractable, we perform inference via Gibbs sampling.%
\footnote{
    Variational inference with pseudopoints in $\theta$-space for the items would be possible for large-scale data.  This would be a relatively simple extension using existing approaches such as \citet{HensmanEtAl:2015}. However, the exact sampling scheme we outline is sufficient for the applications we investigate.
   }
We begin by initializing a Markov chain by sampling initial latent scores $\{\theta\}$ and mean function coefficients $\{\beta\}$ from their respective priors.
Then, given $\{\theta\}$ and $\{\beta\}$, we sample the latent function values corresponding to the observations $\{y_{ij}\}$. Let $\data_j$ represent the responses from respondent $j$ only, $\mathbf{f}_i = \{f_i(\theta_j)\}$ represent the latent function values associated with all responses to item $i$, and $\boldsymbol \theta_i$ be the latent scores of all respondents who answered item $i$.
We sample:
\[
  \mathbf{f}_i \sim \mathcal{N}\bigl(\mu(\boldsymbol \theta_i; \{\beta\}), K(\boldsymbol \theta_i, \boldsymbol \theta_i)\bigr).
\]
Finally, we extend the sampled vectors $\mathbf{f}_i$ to dense samples of each latent function.  This is feasible since the latent space $\Theta$ used in IRT is universally small; in the vast majority of cases, this dimension is 1 or 2. In our implementation, we took a dense grid $\boldsymbol \theta^\ast$ spanning from $-5$ to $5$ in increments of $0.01$, which is sufficient to densely cover the support of the latent score prior. Now for each item $i$, we can sample dense vectors $\mathbf{f}^\ast_i$ on this set from the posteriors induced by $\boldsymbol \theta_i$ and
$\mathbf{f}_i$, which is normal as the posterior on $f_i$ is a Gaussian process:
\begin{align*}
    &\mspace{10mu}p\bigl(\mathbf{f}^\ast_i \mid \boldsymbol \theta^\ast, \boldsymbol \theta, \mathbf{f}_i, \{\beta\}\bigr)
    =
    \mathcal{N}(\mathbf{f}^\ast_i; \mathbf{m}^\ast, \mathbf{C}^\ast),
    \\
    \shortintertext{where}
    \mathbf{m}^\ast &=
    \mu(\boldsymbol\theta^\ast)
    +
    K(\boldsymbol\theta^\ast, \boldsymbol\theta)\, K(\boldsymbol\theta, \boldsymbol\theta)^{-1}\bigl(\mathbf{f}_i -
    \mu(\boldsymbol\theta)\bigr);
    \\
    \mathbf{C}^\ast &=
    K(\boldsymbol\theta^\ast, \boldsymbol\theta^\ast) 
    -
    K(\boldsymbol\theta^\ast, \boldsymbol\theta)
    K(\boldsymbol\theta, \boldsymbol\theta)^{-1}
    K(\boldsymbol\theta, \boldsymbol\theta^\ast),
\end{align*}
and the dependence of $\mu$ on $\{\beta\}$ has been omitted.

With the Markov chain initialized, we proceed as follows. First we sample new latent function values at the observations for each item $\{\mathbf{f}_i\}$, using elliptical slice sampling \citep{ESS}. We then extend the $\{\mathbf{f}\}$ to dense samples 
$\{\mathbf{f}^\ast\}$ as described above. Next we sample each of the latent scores $\theta_j$ from the posterior induced given the latent function samples $\{\mathbf{f}^\ast\}$. Extending these samples onto a dense grid allows us to compute a dense approximation of the \emph{exact} posterior of $\theta_j$ on the grid. The unnormalized posterior is:
\[
  p\bigl(\theta_j \mid \{\mathbf{f}^\ast\}, \data_j\bigr)
  \propto
  p(\theta_j)
  \textstyle\prod_i
  \Pr\bigl(y_{ij} \mid \mathbf{f}^\ast_i(\theta_j) \bigr).
\]
We then use inverse transform sampling to sample a new latent score for each respondent from their posterior.
Finally, we sample new values for the mean function hyperparameters $\{\beta\}$ using a
Metropolis--Hastings step with a Gaussian proposal distribution.
We have implemented this inference method in the R package \texttt{gpirt}. Once this chain has mixed we can then estimate IRFs if desired by pushing a chain of samples of the dense latent functions through the chosen sigmoid and averaging.

The posterior distributions in all IRT models exhibit rotational
invariance where the sign on the latent parameters $\{\theta_j\}$ and the shape of the functions $\{f_i\}$ can be altered to produce an identical likelihood.
In applications, generally either the directionality of
the latent space is not  interesting, in which case
samples from either reflective mode are useful, or the directionality
is known, and posterior samples stuck in a substantively inappropriate mode can be replaced or post-processed by
imposing the desired orientation \citep{Stephens:1997}.

\subsection{ACTIVE LEARNING}

A major benefit of our fully Bayesian model is enabling a direct scheme for adaptive testing via sequential Bayesian experimental design. We consider the following task. Suppose we have estimated IRFs from data $\data$, for the $i$th item estimating:
\[
  \pi_i(\theta^\ast) = \Pr(y_{i}^\ast = 1 \mid \theta^\ast, \data)
\]
by marginalizing the latent function $f_i$. We then seek to adaptively present a series of items to some new respondent so as to learn their latent score as quickly as possible. Here we propose a natural scheme based on maximizing the mutual information between the unknown response to each item and the unknown latent score $\theta^\star$.

Our proposed algorithm works as follows. Let $\data^\star$ represent a dataset  augmenting our training data $\data$ by the (initially empty) available responses from the new respondent.
We initialize our belief about $\theta^\star$ to the chosen prior used during inference: $p(\theta^\star \mid \data^\star) = p(\theta)$.
Now we compute the mutual information between the response $y_{i}^\star$ to item $i$ and $\theta^\star$ given the available data:
\begin{equation}
    I(y_{i}^\star; \theta^\star \mid \data^\star) 
    =
    h(p_{i}^\star)
    -
    \mathbb{E}_{\theta^\star}\bigl[ h(\pi_i(\theta^\star)) \mid \data^\star].
\end{equation}
Here $h(p)$ is the binary entropy function and $p_{i}^\star$ is the \emph{marginal} probability of a positive response to item $i$:
\[
  p_{i}^\star
  =
  \int \pi_i(\theta^\ast)\,p(\theta^\star \mid \data^\star)\,\mathrm{d}\theta^\star.
\]
We present the respondent with the item maximizing the mutual information and augment $\data^\star$ with the response.
We then approximate the updated posterior distribution on $\theta^\star$ by multiplying the current belief by the IRF for the chosen item (for a positive response) or its complement (for a negative response), e.g.\ if $y_{i}^\star = 1$ we take:
\[
  p(\theta^\star \mid \data^\star, y_{i}^\star) \propto
  p(\theta^\star \mid \data^\star) \,\pi_i(\theta^\star).
\]
We repeat this process until a stopping condition is met.
This may be a pre-chosen number of items, but could also be sufficient certainty in $\theta^\star$.
This procedure is both computationally efficient and effective in practice.
Although we could refit the entire GPIRT model each time we get a new response, the extra cost is not likely to be worth it if the training dataset is reasonably sized.

\section{RELATED WORK}

Inference in parametric IRT models is often achieved via some variant of maximum likelihood estimation (MLE),
maximizing \eqref{eq:likelihood} as a function of $\{\beta_i\}$ and $\{\theta_j\}.$
Unfortunately, maximizing the full joint likelihood has proven to be difficult and most parametric models therefore are  estimated using the marginal maximum likelihood (MML) framework \citep{Bock:1981}. 
Here, we assume that the marginal \emph{posterior} distribution of the latent scores $\{\theta_j\}$ is known a priori.
We will denote this assumed marginal distribution $q(\theta)$.
In MML we estimate each respondent's contribution to the likelihood \eqref{eq:likelihood} by marginalizing her latent score under the assumed $\theta$ distribution:
\begin{equation}
\label{eq:MML}
\mathcal{L}_j
\approx
\int
\prod_i
\Pr( y_{ij} \mid \theta_j, f_i )
\,q(\theta_j)\,
\mathrm{d}\theta_j,
\end{equation}
where $\mathcal{L}_j$ represents the component of the likelihood associated with respondent $j$ in \eqref{eq:likelihood}. In this procedure, the item-level parameters are estimated first, marginalizing the latent scores as in \eqref{eq:MML}. 
$\{\theta_j\}$ is then estimated afterwards via some procedure such as calculating the expected a posteriori (EAP) estimate \citep{rizopoulos2006ltm}.

Most nonparametric methods build from Equation~\ref{eq:MML}.
The most relevant approaches here focus on relaxing assumptions about the link function $\sigma$ \citep[but see, e.g.,][]{woods2006item}.
However, problems arise since it is difficult to simultaneously estimate $\{f_i\}$ and $\{\theta_j\}$.  Indeed, the entire idea of the MML approach is to marginalize out $\{\theta_j\}$. Therefore, standard nonparametric IRT models estimate the IRF not based on $\{\theta_j\}$ but instead based on the observed data as a proxy.  

For instance, in kernel-based IRT \citep{Ramsay:1991} we first transform respondents' scores (number correct) into quantiles of a specified latent trait distribution $q(\theta)$.
That is, if respondent $j$ is in the empirical $s_j$th percentile in raw average score across the items, we first estimate their latent score with $\hat{\theta}_j = Q^{-1}(s_j),$
where $Q^{-1}$ is the inverse CDF for $q(\theta)$. Fixing these proxies for the latent scores, we can then use Nadaraya--Watson \citep{nadaraya1964estimating, watson1964smooth} regression to estimate the IRFs by kernel smoothing over the
training data.

Due to the inherent difficulties associated with simultaneously
estimating IRFs and $\{\theta_j\}$, adopting a Bayesian framework
is an attractive option \citep{albert1992bayesian}. Given the
likelihood in Equation~\eqref{eq:likelihood}, all that we need to
complete the model is a prior on $\theta$ and (optionally) on the
parameters defining $\{f_i\}$.  
\citet{albert1993bayesian} used normal priors in the context of the normal ogive model and developed a complete Gibbs sampler for the resulting model.  \citet{ImaiLoOlmstead:2016} estimates this same model using the expectation-maximization (EM) approach, which is the method we use in the Congress application below.  Subsequent work has developed Bayesian versions of nearly all of the common parametric models.

Previous research has also been done in the area of Bayesian nonparametric IRT.
\citet{karabatsos2004order} considered Bayesian inference under the monotone nonparametric framework of Mokken while \citet{arenson2017bayesian} approximated monotone IRFs using a finite mixture of beta distributions.
\citet{duncan2008nonparametric} places a Dirichlet process (DP) prior on the $q(\theta)$ distribution while retaining the 2PL form for the latent functions
and further reported a variant that instead models the IRFs as a DP as well  \citep[see also][]{san2011bayesian}. 
However, we are aware of no existing model that adopts the GP approach we outline above.

The most closely related GP approach to our own might be Gaussian process latent variable models, or GPLVM \citep{Lawrence2004}.
GPLVM is not particularly well-suited for IRT as there is a fundamental mismatch in the choice of likelihood, which is naturally binomial but normal in the GPLVM.
A further issue is that GPLVM typically assumes that all ``dimensions'' (items) have a common/shared error term, which fits poorly in the measurement model domain most closely related to IRT.
Moreover, the mismatch in likelihood makes response prediction less principled than the IRFs provided by GPIRT. Nonetheless, we include GPLVM in our benchmarks.

\section{APPLICATIONS}

We illustrate the benefits of GPIRT with two applications to real-world observational data.
First we embed Members of the US House of Representatives from the 116th Congress (elected in 2018) from their roll-call records.
Here we focus on finding interesting attributes of the IRFs that standard scaling cannot uncover due to their restrictive functional form assumptions.
We then apply the model to a survey dataset, where respondents were given a narcissism battery.
With this data we focus on improving predictive performance, and also illustrate that the model performs strongly in an active learning task.

\subsection{ROLL CALL VOTING IN THE HOUSE OF REPRESENTATIVES}

Members of the U.S.\ House of Representatives give recorded ``yea'' and ``nay'' votes on the various proposals the House considers.
There is a long history in political science of using these votes to embed the legislators in a latent space,  with a one-dimensional left--right ideological continuum being of interest in recent sessions.
The gold standard ideological scores for members of Congress are the DW--NOMINATE scores \citep{Poole:1997} and Bayesian IRT models \citep{Clinton:2004}. Similarly to the 2PL model described above, the NOMINATE procedure assumes a specific (monotonic) functional form for IRFs.

The monotonicity assumption in 2PL and NOMINATE often holds
and these models are highly predictive both in-sample and
out-of-sample. In the Supplementary Material, we show that GPIRT performs equally
well and sometimes better than the 2PL, although prediction is
rarely a task in this setting.  However, the strong parametric
assumptions can obscure important dynamics in key roll-call votes,
which can in turn result in embeddings that correspond poorly with
other evidence about members' ideology.

Our focus in this
application, therefore, is on recovering interesting aspects of IRFs
and the qualitative value of the resulting embeddings.  For example,
in the 2019 session several proposals drew ``nay'' votes from both
ends of the ideological spectrum, as Republicans voted against
bills for being too liberal, and liberal Democrats voted
against bills for not being liberal enough. In such a case, standard
parametric assumptions result in extreme members to
instead be scaled as moderates.  Indeed, DW--NOMINATE rates Rep.\
Alexandria Ocasio--Cortez, a member of the Democratic Socialists of
America, as more conservative than 86\% of Democrats in the House of
Representatives \citep{citeXu:2006}.  The procedure similarly scales
the other members of ``The Squad,'' Reps.\ Ilhan Omar, Ayanna
Pressley, and Rashida Tlaib, as among the most conservative Democrats
in the chamber.  The more flexible approach of
the GPIRT model does not force extreme members who vote against
their party to be scaled as moderates.

We estimate GPIRT latent traits and IRFs using all roll call votes in
the first session of the U.S.\ House of Representatives for the 116th
Congress.  We use only votes where the minority vote is at least 1\% of the total with a quadratic mean function as we anticipate some non-monotonic IRFs.
We include all members of the House who voted on at least one bill, with the exception of Rep.\ Justin Amash who left the Republican party.

\begin{figure}
    \centering
    \includegraphics[width=0.9\linewidth]{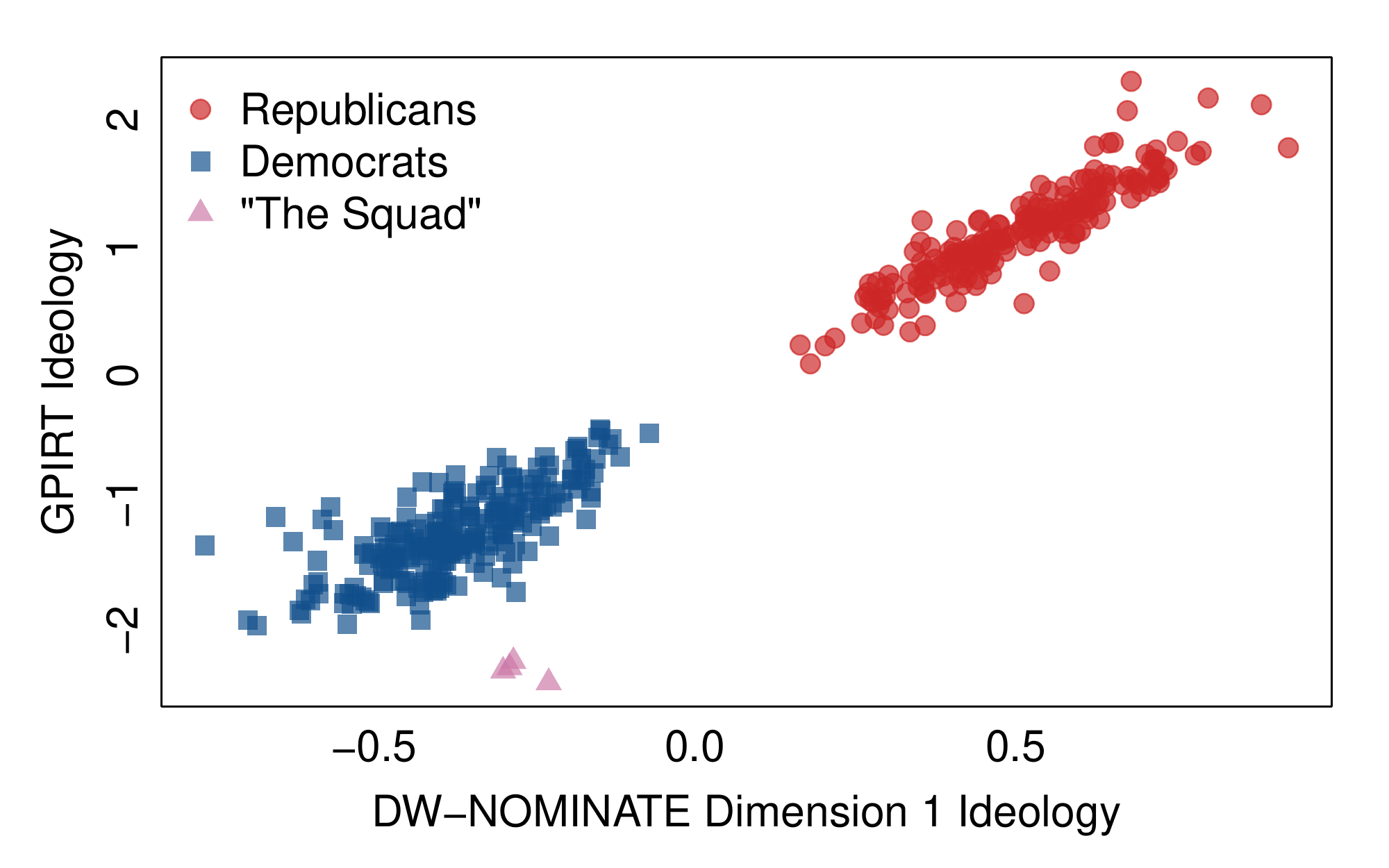}
    \caption{\footnotesize Ideology estimates for members of the 116th House of Representatives for the GPIRT model and DW--NOMINATE.
             Republicans' estimates are plotted in red circles and Democrats' estimates are plotted in blue squares,
             except Reps.\ Ocasio--Cortez, Omar, Pressley, and Tlaib (popularly known as ``The Squad''), who
             are noted with purple triangles.}
    \label{fig:gpirt-vs-nominate}
\end{figure}

Figure~\ref{fig:gpirt-vs-nominate} compares GPIRT ideology estimates with their DW--NOMINATE scores.
We can see that GPIRT and DW--NOMINATE largely agree on the relative ideological placement of members. However, GPIRT finds The Squad to be the most liberal members of the House as we would substantively expect.

\begin{figure*}[ht!]
    \centering
    \begin{subfigure}[b]{0.31\textwidth}
        \includegraphics[width=\textwidth]{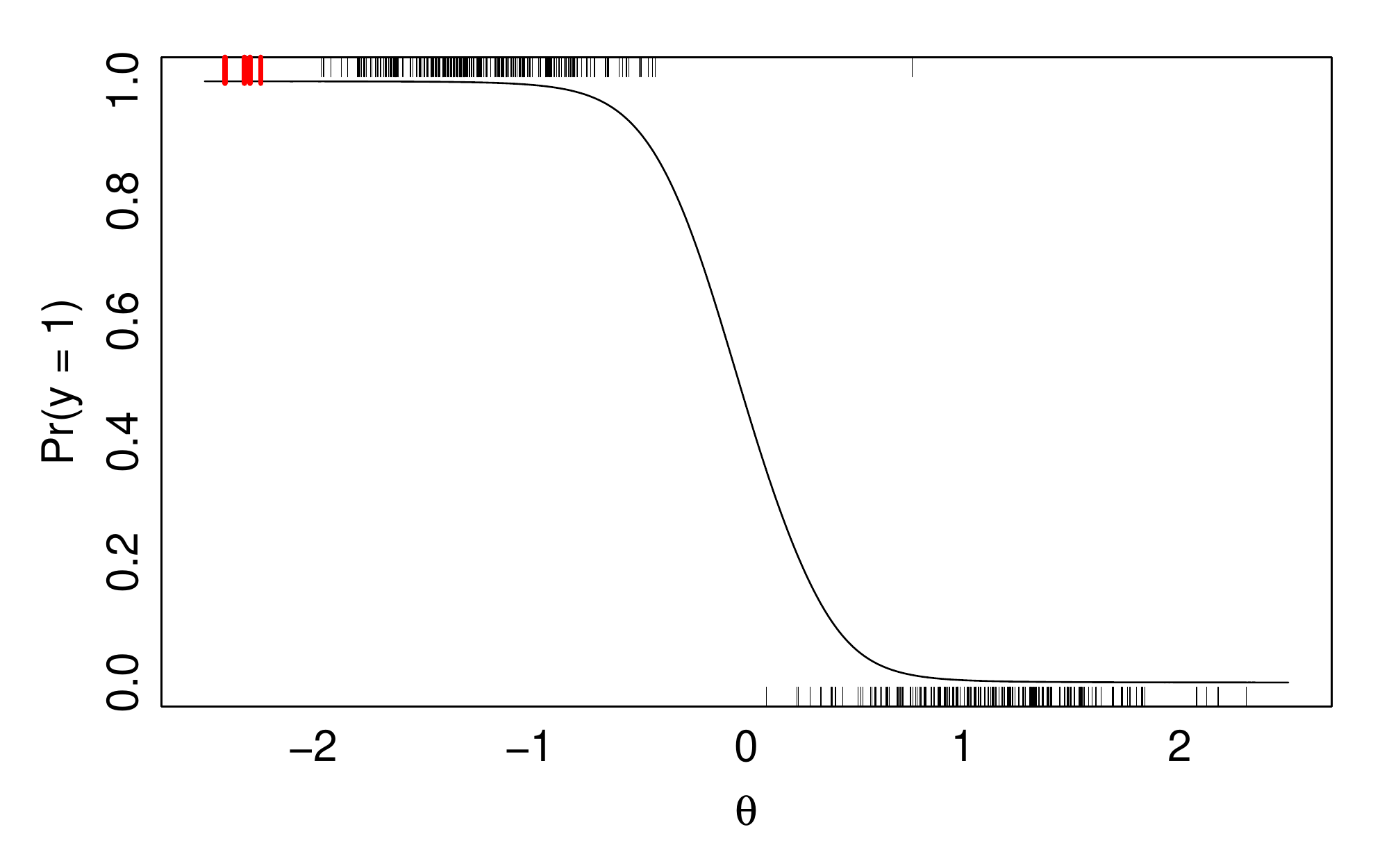}
        \caption{IRF for HR2722, the ``Securing America's Federal Elections Act''}
        \label{fig:safe-act-irf}
    \end{subfigure}
    \hspace{0.02\textwidth}
    \begin{subfigure}[b]{0.31\textwidth}
        \includegraphics[width=\textwidth]{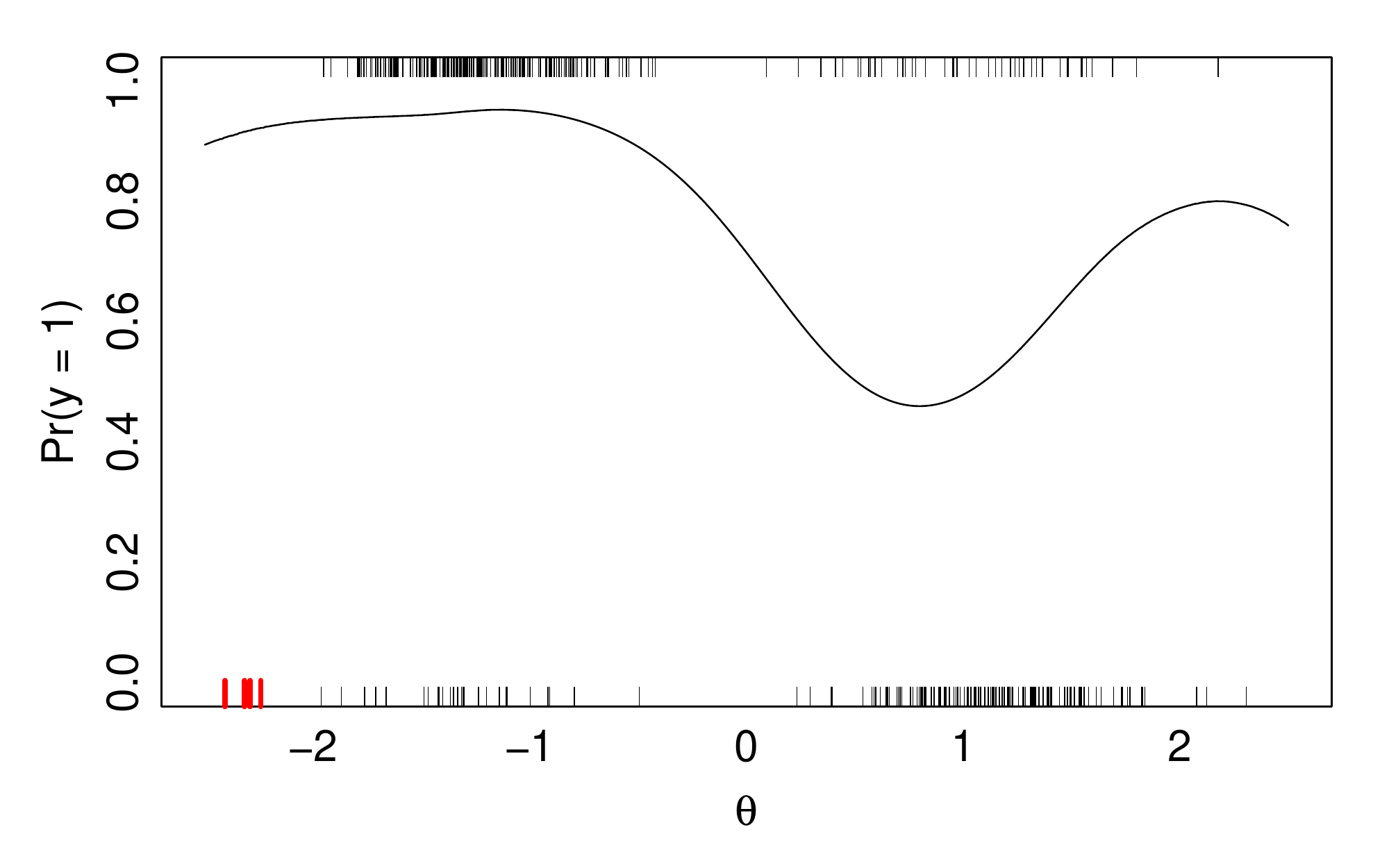}
        \caption{IRF for an amendment to HR2500 ending the Cyprus arms embargo}
        \label{fig:cyprus-irf}
    \end{subfigure}
    \hspace{0.02\textwidth}
    \begin{subfigure}[b]{0.31\textwidth}
        \includegraphics[width=\textwidth]{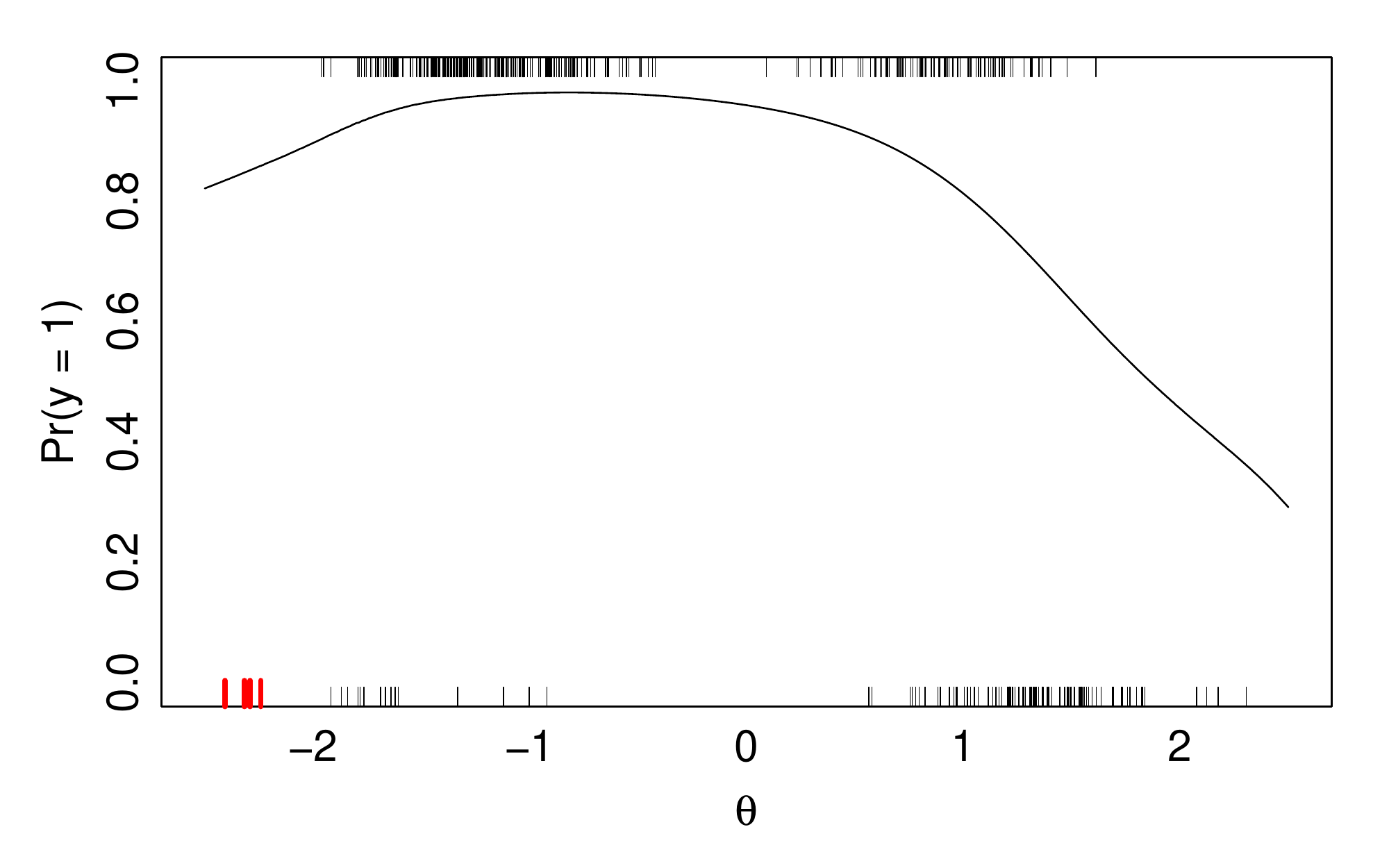}
        \caption{IRF for HJRES31, a budgetary resolution to avoid a government shutdown}
        \label{fig:budget-res-irf}
    \end{subfigure}
    \begin{subfigure}[b]{0.31\textwidth}
        \includegraphics[width=\textwidth]{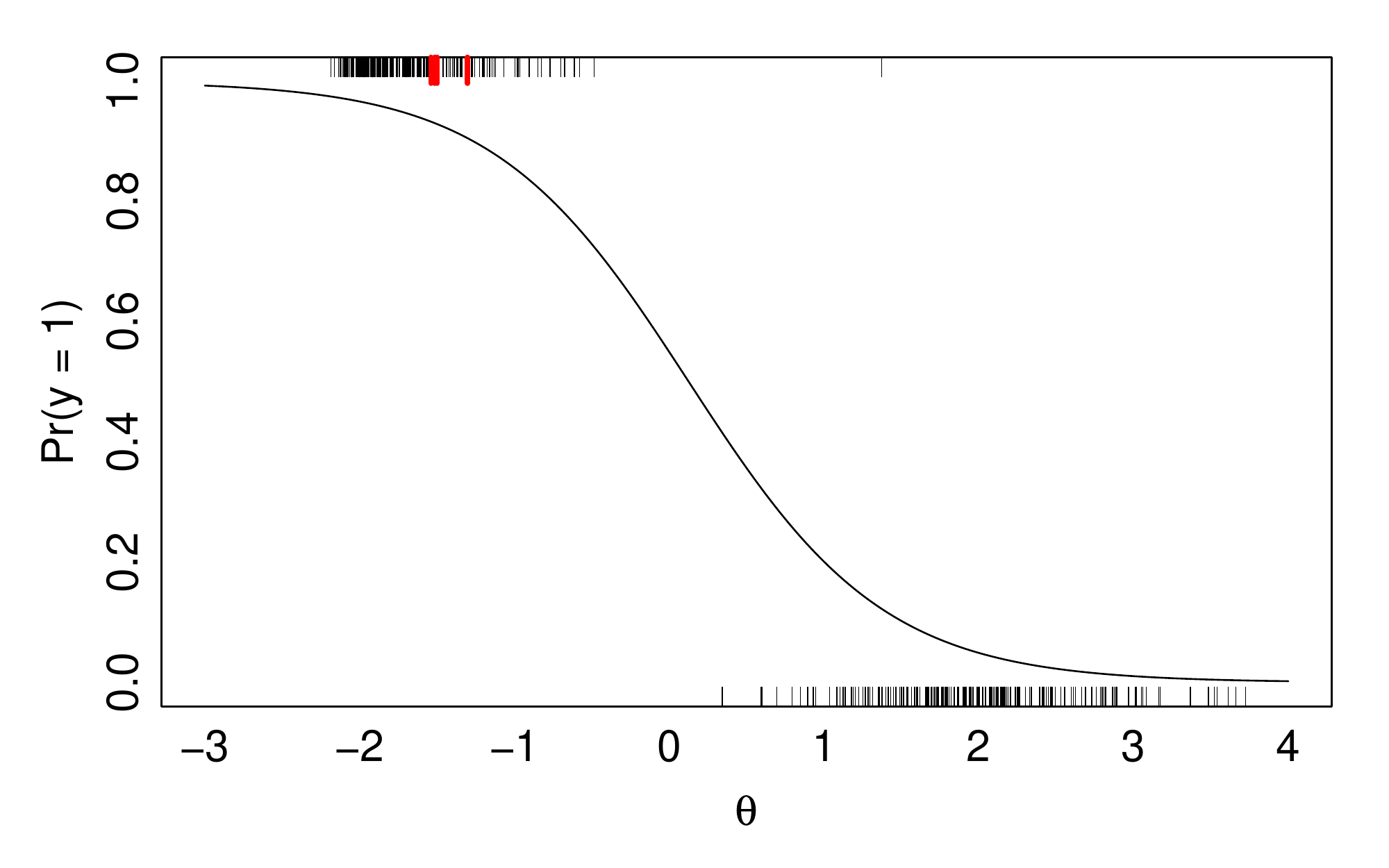}
        \caption{2PL IRF for HR2722}
        \label{fig:safe-act-irf-2pl}
    \end{subfigure}
    \hspace{0.02\textwidth}
    \begin{subfigure}[b]{0.31\textwidth}
        \includegraphics[width=\textwidth]{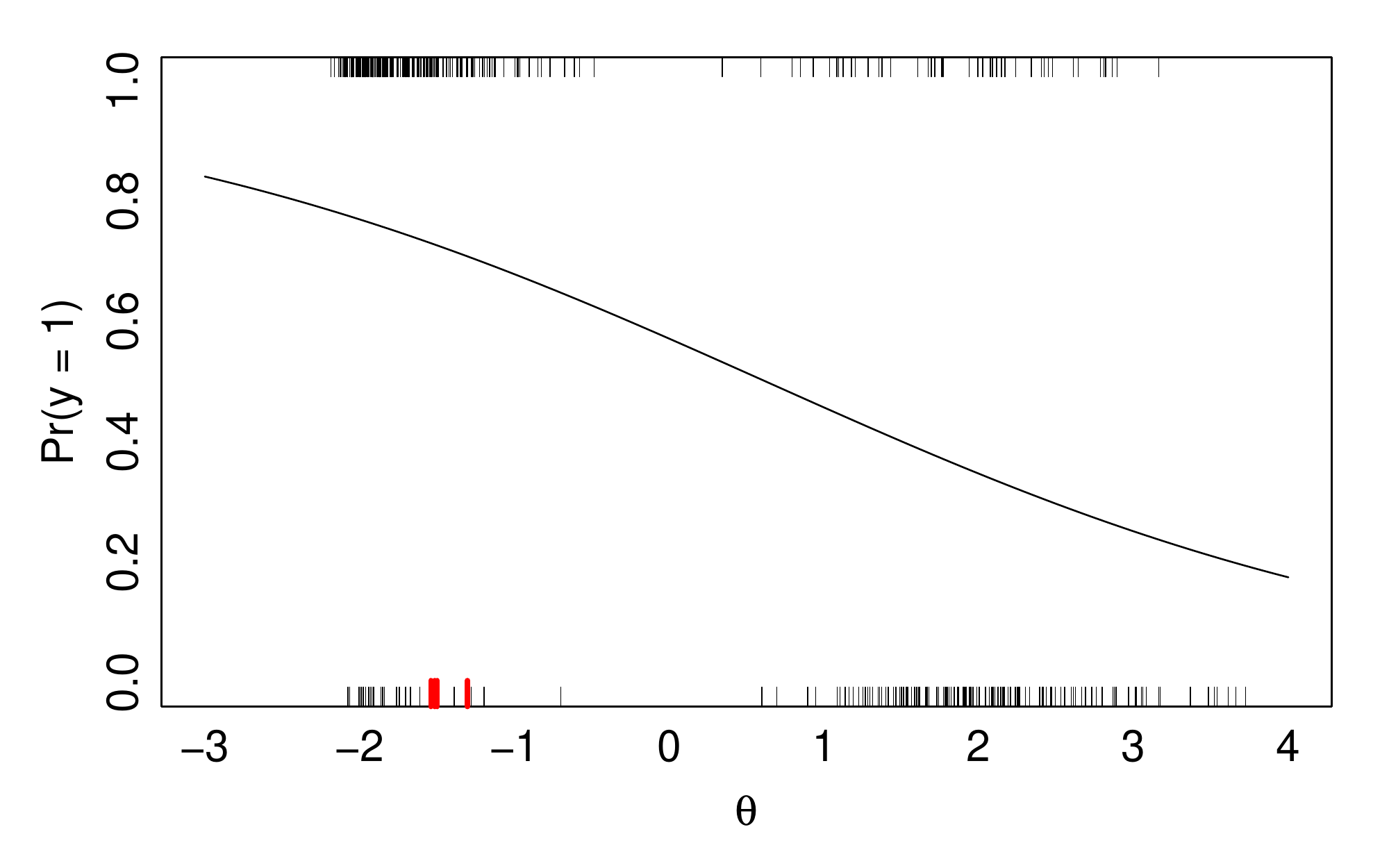}
        \caption{2PL IRF for the Cyprus amendment}
        \label{fig:cyprus-irf-2pl}
    \end{subfigure}
    \hspace{0.02\textwidth}
    \begin{subfigure}[b]{0.31\textwidth}
        \includegraphics[width=\textwidth]{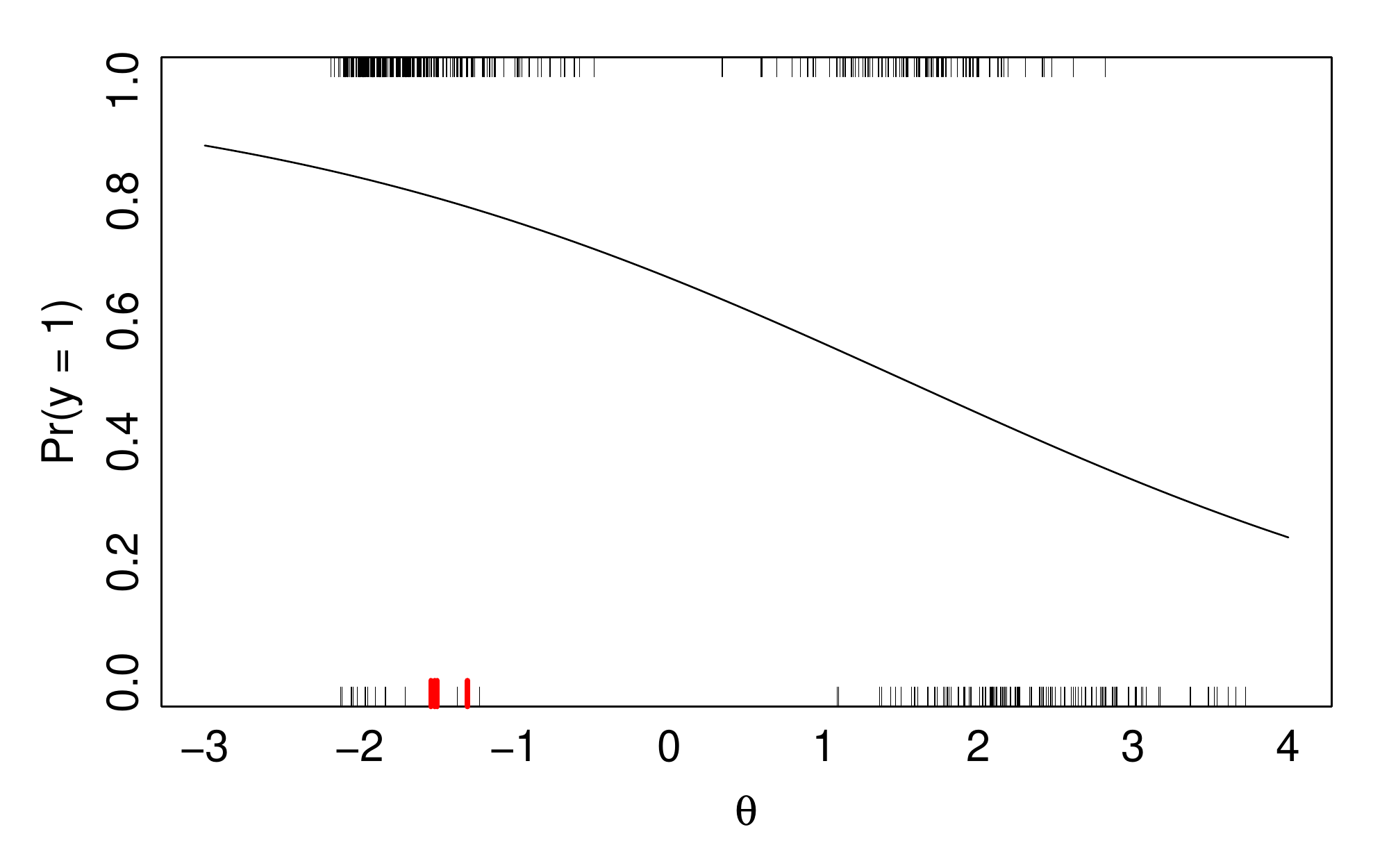}
        \caption{2PL IRF for HJRES31}
        \label{fig:budget-res-irf-2pl}
    \end{subfigure}
    \caption{\footnotesize Example IRFs for three roll call votes in the U.S. House of Representatives' first session of the 116th Congress.
             $\theta$ estimates for members who voted ``Yea'' are displayed in a rug at the top of the plots,
             while $\theta$ estimates for members who voted ``Nay'' are displayed in a rug at the bottom.
             Rug lines for members of the Squad are drawn in red.}
    \label{fig:gpirt-vs-bayesian-irt-irfs}
\end{figure*}

For most roll calls we observe the linear, monotonic patterns that
parametric IRT-like models require (Figure 3a). However, we also observe other patterns, such as the
non-monotonic IRFs depicted in Figures~\ref{fig:cyprus-irf}
and~\ref{fig:budget-res-irf}. Here, the traditional IRT model treats this as uninformative vote with a
relatively flat IRF (Figure~\ref{fig:cyprus-irf-2pl}), while the GPIRT detects the more nuanced ideological structure of the vote including non-saturation,
non-monotonicity, and asymmetry in $\theta$  (Figure~\ref{fig:cyprus-irf}).

Sometimes the ability to account for non-monotonic voting patterns is necessary to catch an important legislative dynamic. For an example, consider the item response functions for H.J.\ Res.\ 31, a funding bill to end a partial government shutdown, depicted in Figures~\ref{fig:budget-res-irf} and~\ref{fig:budget-res-irf-2pl}.
Here, the vast majority of Democrats supported the bill along with many moderate Republicans.
Conservative Republicans opposed the bill on the grounds that it did not include funding for the border wall, while liberal Democrats such as Ocasio--Cortez, Omar, Pressley, and Tlaib argued it did not sufficiently reduce funding for border detention facilities \citep{rollcall-appropriations}.
The GPIRT is able to recover this non-monoticity as shown in Figure~\ref{fig:budget-res-irf}, while the 2PL IRF shown in Figure~\ref{fig:budget-res-irf-2pl} treats the item as largely not informative.

\subsection{NARCISSISTIC PERSONALITY INVENTORY}

We also apply the GPIRT model to responses to a 40 item narcissistic personality inventory (NPI) \citep{RaskinTerry:1988}, which measures one's ``grandiose yet fragile sense of self and entitlement as well as a preoccupation with success and demands for admiration'' \citep[pp.\ 440--441]{Ames:2006}. For each item on the NPI, respondents are presented with two statements and asked to select which one fits them best; an example item is ``Modesty doesn't become me'' (positive) vs.\ ``I am essentially a modest person'' (negative).
Responses to the complete inventory were collected  from the Open
Source Psychometrics Project \citep{OpenPsychometrics}  ($n= 10,440$)
and a convenience sample from the Qualtrics panel ($n=2,945$).
We estimate the 2PL IRT, kernel-smoothed IRT, GPLVM, and GPIRT models on a randomly selected sub-sample of 2,000
respondents.
The IRFs in the GPIRT showed  deviations from the standard 2PL IRT model for several items.

Permitting flexibility in the IRFs allows the GPIRT to perform better than the comparison models as measured by predictive performance on held-out responses.
To show this, we performed the following experiment.
We randomly selected 20\% of the available observations and treated them as missing, then estimated the IRFs and the respondents' latent traits from the remaining data.
We then used the trained model to predict the missing responses and calculated the log likelihood, predictive accuracy, and area under the receiver operating characteristic curve (AUC) for the the held out observations using the trained model; we repeated this procedure 20 times.
The mean log likelihood for the held out observations is reported in Table~\ref{tab:npi-held-out}.  

\begin{table}
    \footnotesize
    \caption{
        Fit for GPIRT, 2PL, GPLVM, and kernel-smoothed IRT models on NPI responses with 20\% of the data held out.
        $\varepsilon \approx 2.2\times10^{-16}$ is machine epsilon.
    } 
    \label{tab:npi-held-out}
    \begin{center}
        \begin{tabular}{lclcc}
            \toprule
            \multicolumn{1}{c}{Model}
            & $\mathcal{L}/N$
            & $t$-test vs. GPIRT
            & AUC
            & Accuracy \\
            \midrule
            GPIRT   & -0.52 &                                               & 0.82  & 0.74  \\
            2PL     & -0.73 & 0.20 $\left(p < \varepsilon\right)$           & 0.66  & 0.63  \\
            GPVLM   & -1.39 & 0.87 $\left(p < \varepsilon\right)$           & 0.71  & 0.66  \\
            KS IRT  & -0.62 & 0.10 $\left(p \approx 10\varepsilon\right)$   & 0.68  & 0.68  \\ 
            \bottomrule
        \end{tabular}
    \end{center}
\end{table}

For the NPI dataset, the GPIRT outperformed the comparison models,
with a higher mean log likelihood than any model in the comparison set.
A paired \(t\)-test confirms these are significant improvements in model fit, and is also confirmed by comparing the mean accuracy and AUC across the 20 simulations.

\subsection{ACTIVE LEARNING}

We also used these datasets to evaluate our adaptive testing procedure. 
For the NPI dataset, we compare our procedure to a published reduced-form NPI battery \citep{Ames:2006}.
The reduced-form battery contains 16 questions deemed by experts to be a suitable subset.

For 1,000 randomly selected respondents that were not included in our initial training sample, we estimated their latent traits using their actual responses to the full 40-item battery and our estimated GPIRT IRFs.
For each respondent, we then estimated their latent trait using the 16-item reduced-form battery from \citet{Ames:2006}, 16 items chosen using our adaptive testing scheme, and 16 randomly selected items.
We calculated the root mean squared error (RMSE) of these batteries' $\theta$ estimates with the full-battery estimates; the results are presented in Table~\ref{tab:cat-rmse}.
Notably the RMSE for the adaptive battery gives a 20\% improvement over randomly selected items,
whereas the reduced-form battery actually performs \emph{worse} than random selection.

\begin{table}
    \centering
 \footnotesize
    \caption{RMSE for 1000 responses to 16 items} 
    \label{tab:cat-rmse}
    \begin{tabular}{rrrr}
        \toprule
        & CAT & Fixed & Random \\ 
        \midrule
                             RMSE & 0.257 & 0.338 & 0.321 \\ 
        Improvement vs.\ random & 20\% & $-5.5\%$ & --- \\ 
        \bottomrule
    \end{tabular}
\end{table}

\section{CONCLUSION}

In this article we provided a fully Bayesian nonparametric IRT model
that allows for the simultaneous estimation of ability parameters and
IRFs while allowing for high levels of flexibility in the IRF shapes.
We showed that this model performs better than standard parametric
models in terms of estimating unusual IRFs, predictive accuracy, and in an active learning setting. 

Trivial extensions include allowing for multiple dimensions and categorical response functions.
Additionally, we could avoid the independence assumption on the items and couple them together via a multi-task kernel.
For example if the items were parameterized by some vector $x$ we could take a product kernel such as $K(x, x')K(\theta, \theta')$.
While a strength of the method is avoiding potentially unfounded assumptions about monotonicity, saturation, and symmetry, we could impose monotonicity via EP to impose derivative constraints \citep[e.g.][]{RiihimakiVehtari10}, saturation via a slightly modified likelihood, and symmetry via an appropriately modified kernel (e.g. it would suffice to take $K(|x - c|, |x' - c|)$, where $K$ is a desired base kernel, to impose symmetry around $c$).
Other potential extensions are to include feature vectors in the learning kernel to allow, for instance, smooth changes in ability over time or differential item functioning for subgroups.  

\subsubsection*{Acknowledgements}

RG was supported by the National Science Foundation (NSF) under award numbers IIS--1939677, OAC--1940224, and IIS--1845434. JM was supported by the NSF under award number SES--1558907.

\bibliography{gpirt-references}
\bibliographystyle{icml2020}

\clearpage
\newpage
\appendix
\begin{center}
    \textbf{\Large SUPPLEMENTARY MATERIAL}
\end{center}

\setcounter{table}{0}
\renewcommand{\thetable}{A\arabic{table}}

\section{SUPPLEMENTAL RESULTS FOR THE US CONGRESS APPLICATION}
\label{sec:supplemental-congress-results}

The US Congress application presented in the main paper focused on improved substantive insights about the meaningful embedding of legislators in the latent space and the shape of specific IRFs. For completeness, we also assessed (i) out-of-sample performance and (ii) our active learning procedure.  

The out of sample results are given in Table~\ref{tab:held-out}. Indeed, the paired \(t\)-test on the log likelihood shows no significant difference. 

\begin{table}[h]
    \begin{center}
        \caption{Fit for GPIRT and the 2PL models with data held out.} 
        \label{tab:held-out}
        \begin{tabular}{cp{1in}rrrr}
            \toprule
            & & \multicolumn{2}{c}{Congress} \\
            \cmidrule{3-4} \cmidrule(lr){5-6}
            \multicolumn{1}{c}{Hold out} &
            \multicolumn{1}{c}{Statistic} &
            \multicolumn{1}{c}{GPIRT} &
            \multicolumn{1}{c}{2PL} \\ 
            \midrule
            \multirow{6}{*}{10\%}   & Mean $\mathcal{L}$        &  $-3208$ & $-3201$                \\
                                    & Mean $\mathcal{L}$ / N    & $-0.123$ & $-0.122$               \\
                                    & Diff. in means            & \multicolumn{2}{c}{$-0.0003$}     \\
                                    & (paired $t$-test)         & \multicolumn{2}{c}{($p = 0.53$)}  \\
                                    & Mean accuracy             & $0.957$   & $0.953$               \\
                                    & Mean AUC                  & $0.991$   & $0.990$               \\[0.5em]
            \multirow{6}{*}{20\%}   & Mean $\mathcal{L}$        & $-6466$   & $-6443$               \\ 
                                    & Mean $\mathcal{L}$ / N    & $-0.124$  & $-0.123$              \\
                                    & Diff. in means            & \multicolumn{2}{c}{$-0.0004$}     \\
                                    & (paired $t$-test)         & \multicolumn{2}{c}{($p = 0.09$)}  \\
                                    & Mean accuracy             & $0.957$   & $0.953$               \\
                                    & Mean AUC                  & $0.991$   & $0.990$               \\
            \bottomrule
        \end{tabular}
    \end{center}
    \footnotesize
    Mean  $\mathcal{L}$ is the average log-likelihood across the 20 replications where the held out sample is 10\% or 20\%.
    Mean $\mathcal{L}$ / N is the mean  $\mathcal{L}$ divided by the number of responses held out. 
    Mean accuracy  is the proportion correctly classified of the held out responses.
\end{table}

In an experiment with the House of Representatives data, we held out 20\% of the members and estimated IRFs for the non-unanimous items given those missing members.
We then estimated those held out members' latent trait using all estimated IRFs and the members' actual responses, and compared the estimate to that obtained using the members' responses to 10 randomly selected roll call votes and 10 roll calls selected using our adaptive testing scheme.
The RMSE for the adaptive estimates were a 16.97\% improvement over the estimates from the randomly chosen items.

\section{FULL STATEMENT OF THE MODEL}
\label{sec:model-graph}

The joint probability of \( \left( \mathbf{y}, \{\mathbf{f}_i\}, \boldsymbol\beta, \boldsymbol\theta \right) \) is
\begin{multline*}
    \pi \left( \boldsymbol\theta, \{\mathbf{f}_i\} \mid \mathbf{y} \right)
    \triangleq \\
    \mspace{6mu}g\left(\boldsymbol\theta; \mu_\theta, \Sigma_\theta\right)
    g\left(\boldsymbol\beta; \mu_\beta, \Sigma_\beta\right) \times {} \dotsm \\
    \prod_i g\left( \mathbf{f}_i; \mu\left(\boldsymbol\beta_i, \boldsymbol\theta\right), K\left(\boldsymbol\theta; \sigma_f, \ell\vphantom{_f}\right) \right) \times {} \dotsm \\
    \prod_i \prod_j \left[ \dfrac{1}{1 + \exp\left(-\mathbf{y}_{ij}\mathbf{f}_{ij}\right)} \right],
\end{multline*}
where $g$ is the multivariate normal p.d.f.,
$\mu$ is the chosen mean function,
and $K$ is the squared exponential covariance function.
$\mu_\theta$ and $\Sigma_\theta$ are the mean and covariance of the prior on $\boldsymbol\theta$,
which we fix to $\mathbf{0}$ and $\mathbf{I}$ (a standard normal).
Similarly, $\mu_\beta$ and $\Sigma_\beta$ are the mean and covariance of the prior on $\boldsymbol\beta$.
$\sigma_f$ and $\ell\vphantom{_f}$ are the scale factor and length scale of $K$ respectively;
we also fix those, each to $1$.

\end{document}